\documentclass[conference]{IEEEtran}

\usepackage{xspace}
\usepackage{graphicx}
\usepackage{caption}
\usepackage{booktabs}
\usepackage{makecell}
\usepackage{algorithm}
\usepackage{algorithmic}
\usepackage{amsmath}
\usepackage{amssymb}
\usepackage{multirow}
\usepackage{capt-of}
\usepackage[capitalize]{cleveref}
\IEEEoverridecommandlockouts
\newcommand{\headercell}[1]{%
  \begin{tabular}[c]{@{}c@{}}
    #1
  \end{tabular}%
  \hspace{0.15em}$\uparrow$%
}

\newcommand{\method}{FedE2E\xspace}

\title{FedE2E: One-Shot Federated Learning for End-to-End Autonomous Driving with Closed-Loop Evaluation}

\author{
    \IEEEauthorblockN{Tao Lian\thanks{This work was supported by the Spanish grant Ref. PID2024-157936NB-I00 (HAMILTON project), funded by MICIU/AEI/10.13039/501100011033 and by FEDER, EU. A. M. López acknowledges financial support for his general research activities from ICREA under the ICREA Academia Program. T. Lian acknowledges support for his research from the China Scholarship Council through the Ph.D. Grant Ref. 202102890072. All authors acknowledge the support of the Generalitat de Catalunya through the CERCA Program and its ACCIÓ Agency for CVC’s general activities.}}
    \IEEEauthorblockA{\textit{Computer Vision Center (CVC)} \\
    \textit{Univ. Autònoma de Barcelona (UAB)}\\
    Barcelona, Spain \\
    tlian@cvc.uab.cat}
    \and
    \IEEEauthorblockN{Antonio M. L\'opez}
    \IEEEauthorblockA{\textit{Computer Vision Center (CVC)} \\
    \textit{Univ. Autònoma de Barcelona (UAB)}\\
    Barcelona, Spain \\
    antonio@cvc.uab.cat}
}
\begin{document}
\maketitle
\begin{abstract}
Federated learning enables collaborative model training without sharing datasets. However, existing federated learning methods for autonomous driving typically require multi-round communication and lack closed-loop evaluation, while one-shot methods mainly focus on perception tasks. To address these limitations, we propose \method, a one-shot federated learning method for end-to-end autonomous driving. It comprises multi-client calibration, anchor-based feature alignment, and token-wise consensus distillation to distill knowledge from heterogeneous models. Closed-loop evaluation on Bench2Drive demonstrates the effectiveness of \method in both one-shot and multi-round federated learning. In the one-shot setting, it achieves a Driving Score of 55.85, which is 4.07 points higher than prior work. In the multi-round setting, it reaches 60.35 after five communication rounds, the best performance among all compared methods under the same number of communication rounds.

\end{abstract}    
\section{Introduction}
\label{sec:intro}
End-to-end driving is a key approach to autonomous driving, which directly maps sensor data, vehicle states, and navigation commands to control actions or trajectories, simplifying traditional modular architectures ~\cite{chen2024end,codevilla2018end}. However, these models require substantial driving data, which cannot be easily shared across countries or institutions due to privacy concerns. Consequently, models trained by individual organizations often cover a limited range of driving scenarios and generalize poorly across cities.

Federated learning enables multiple clients, such as mobile devices, vehicles, companies, and research institutions, to collaboratively train a global model by uploading model weights or gradients without sharing local data, thereby reducing privacy concerns \cite{mcmahan2017communication, chellapandi2024federated, kalloori2024crosssilo}. Although existing methods have demonstrated promising results in image classification \cite{yu2024contrastive,chen2023federated, zhang2023federated,sun2023feature}, they typically require multiple rounds of client-server communication, limiting their applicability when only pre-trained client models are available. For example, organizations may share trained model weights once without participating in multi-round training. One-shot federated learning, which requires only a single model upload from each client, offers a practical alternative. However, existing one-shot methods mainly focus on classification tasks, with limited exploration in end-to-end autonomous driving \cite{zhang2022dense,heinbaugh2023datafree, yang2024feddeo,lian2026feds2r}. Moreover, existing multi-round federated learning approaches designed for end-to-end autonomous driving are often evaluated using open-loop metrics such as root mean squared error (RMSE) and mean absolute error (MAE)
\cite{zhang2021endtoend,zhang2021realtime,
nguyen2022deep,aparna2021steering}. These metrics measure prediction errors on specific datasets but cannot capture error accumulation during continuous vehicle-environment interactions. Therefore, improved open-loop accuracy does not necessarily lead to better performance in real-world driving. Although a few studies adopt closed-loop evaluation \cite{kou2023communication}, their training and testing scenarios remain on the same map, making it difficult to evaluate generalization across unseen driving environments.

\begin{figure*}[t]
    \centering
    \includegraphics[width=1\textwidth]{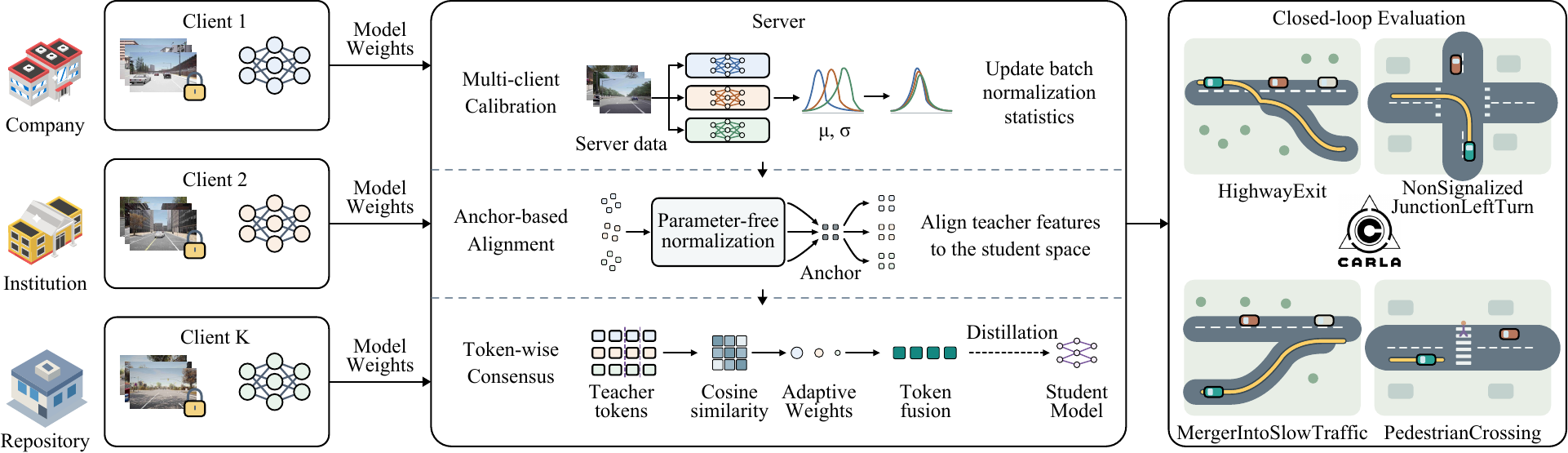}
    \caption{Overview of the proposed FedE2E framework. In our setting, clients can be companies, research institutions, or model repositories that independently train models on their local datasets and upload only model weights to the server once, typically without participating in subsequent training. Once the client models are uploaded to the server, our framework distills knowledge from these models through multi-client calibration, anchor-based alignment, and token-wise consensus distillation to obtain a global model, which is subsequently evaluated in the CARLA simulator for closed-loop evaluation.}
    \label{fig:framework}
\end{figure*}
To address these challenges, we propose \method, a one-shot federated learning framework for end-to-end autonomous driving. Without accessing local data, the server uses client models as teachers to transfer their driving knowledge to a student model (global model) via knowledge distillation. As illustrated in \cref{fig:framework}, \method distills knowledge from multiple client models through three components: multi-client calibration, anchor-based alignment, and token-wise consensus distillation. Specifically, multi-client calibration adapts the batch normalization (BN) \cite{ioffe2015batch} statistics of each teacher model to the server-side dataset domain, while anchor-based alignment maps teacher features into the student representation space. Token-wise consensus distillation then assigns adaptive weights based on feature consistency across teachers and fuses their aligned features to construct distillation targets for the student model. Unlike conventional federated learning methods that require multiple rounds of communication \cite{mcmahan2017communication,li2020federated,lin2020ensemble}, \method enables server-side knowledge distillation from client models without further access to client data or client training. This setting is particularly relevant when organizations can only share their trained models rather than participate in continuous federated training. We conduct closed-loop evaluation in the CARLA simulator \cite{dosovitskiy2017carla} using the Bench2Drive benchmark \cite{jia2024bench2drive} and compare it with representative methods under the same experimental settings. To the best of my knowledge, we are the first to propose a one-shot federated learning framework for end-to-end autonomous driving with closed-loop evaluation. In the one-shot setting, \method achieves Driving Scores (DS) of 55.85 and Success Rates (SR) of 24.7\%, outperforming the previous best method by 4 points in both DS and SR. We also evaluate our method in the multi-round setting, where it achieves 60.35 DS, an improvement of 1.58 points. These results demonstrate the effectiveness of multi-client feature alignment and knowledge distillation in improving closed-loop driving performance and communication efficiency. Our main contributions are summarized as follows:
\begin{enumerate}
    \item We propose a one-shot federated learning framework for end-to-end autonomous driving that requires clients to upload their model weights only once, without sharing local data.
    \item We not only reduce communication to reach comparable driving performance but also achieve the best performance with a single communication round.
    \item We validate the effectiveness of one-shot federated learning for end-to-end driving across different training and testing maps in the CARLA simulator.
\end{enumerate}
\section{Related work}
\label{related work}
\noindent\textbf{Federated learning}. Federated learning has been applied to autonomous driving tasks, including steering angle prediction, trajectory prediction, and vehicle control. Zhang et al. \cite{zhang2021endtoend} applied federated learning to steering angle prediction through multiple rounds of local training and model aggregation. They later proposed a real-time asynchronous federated learning method \cite{zhang2021realtime} to reduce synchronization delays caused by device heterogeneity. However, these methods still require multiple rounds and rely on RMSE for open-loop evaluation, without validating closed-loop driving performance. Aparna et al. \cite{aparna2021steering} investigated steering angle prediction in vehicle-to-everything (V2X) networks using an open-loop MSE evaluation, in which vehicles exchange model weights rather than images. Nguyen et al. \cite{nguyen2022deep} proposed a decentralized federated learning method that enables peer-to-peer model exchange and aggregation, but evaluates performance only using open-loop RMSE. Kou et al. \cite{kou2023communication} introduced CRCHFL, a hierarchical federated learning framework that optimizes communication resources across vehicles and servers. It further demonstrates closed-loop driving in CARLA through qualitative comparisons in Town01 and Town02. However, its evaluation mainly focuses on selected driving scenarios, without comprehensive closed-loop metrics or cross-map generalization results. Therefore, its generalization to unseen maps and complex driving scenarios is not validated. We focus not only on a single round of communication but also on cross-map generalization under closed-loop evaluation.

\noindent\textbf{Feature Alignment}. Feature alignment plays an important role in transferring knowledge across different neural networks. Tung and Mori \cite{tung2019similarity} proposed similarity-preserving knowledge distillation to transfer pairwise feature relationships without directly matching teacher and student representations. Chen et al. \cite{chen2021cross} introduced semantic calibration to adaptively match intermediate layers between teacher and student networks. More recently, Achara et al. \cite{achara2026multiway} investigated multi-model representation alignment and addressed the quadratic number of mappings required by pairwise alignment through a shared representation space. However, these methods mainly focus on representation alignment without explicitly addressing the mismatch between client-specific BN statistics and server-side distillation data. In contrast, our method first calibrates the BN statistics of independently trained client models using a shared server-side dataset, and then performs student-anchored Procrustes alignment. Unlike approaches that optimize alignment mappings during distillation, our alignment matrices are computed offline using singular value decomposition (SVD) and remain fixed throughout student training.

\noindent\textbf{Multi-teacher knowledge distillation.} Liu et al. \cite{liu2020adaptive} proposed adaptive instance-level teacher weighting to combine knowledge from multiple teachers. In federated learning, Lin et al. \cite{lin2020ensemble} proposed FedDF, which uses server-side unlabeled data to distill knowledge from the averaged predictions of multiple client models. However, FedDF performs output-level distillation without explicitly addressing differences in feature representations among independently trained client models. Gong and Wen \cite{gong2024bilevel} introduced BOMD, which uses orthogonal projections to align teacher features with the student feature space and bi-level optimization to learn teacher weights. More recently, Iordache et al. \cite{iordache2025multilevel} fused intermediate features from teachers trained on distinct datasets through a jointly fine-tuned teacher architecture. These approaches demonstrate the benefits of adaptive knowledge fusion, but their weighting strategies do not explicitly account for differences in teacher contributions across local feature representations. In contrast, our token-wise consensus distillation assigns adaptive weights based on feature consistency at each token and combines them with uniform weights to preserve complementary knowledge, without introducing additional learnable weighting parameters.
\section{Method}
\label{method}
In this section, we present \method, designed for end-to-end autonomous driving, particularly in extreme conditions where the client cannot participate in retraining.  \method mainly consists of three parts: (1) Multi-client calibration to mitigate domain shifts caused by different datasets; (2) Anchor-based alignment to reduce feature inconsistency in cross-client representation spaces; and (3) Token-wise consensus distillation, which assigns adaptive weights based on feature consistency among client models and distills multi-client knowledge to a global model. \method requires only one round of communication between the server and the client, thereby reducing subsequent client interactions.

\subsection{Problem Definition}
Given $K$ clients and a server, each client has its own local dataset $\mathcal{D}_k=\{(x_k^i,y_k^i)\}_{i=1}^{N_k}$ that cannot be shared with the server or other clients. $x_k^i$ includes sensor data from the ego-vehicle, vehicle status, and high-level navigation commands; $y_k^i$ denotes the corresponding driving actions, including steering angle, brake, and throttle.
Each client uses its own local dataset to train a client model $\theta_k$ with the same architecture. The server also has its own dataset
$\mathcal{D}_s=\{(x_s^i,y_s^i)\}_{i=1}^{N}$, which has the same data structure as the client datasets. Note that all clients have no access
to the server dataset. The goal is to train a global model $\theta_s$
using the client model weights and the server dataset without access to the client datasets. Each client uploads its model weights only once
and does not participate in subsequent server-side training. Our objective is to obtain a global model that outperforms individual client models and approaches the performance of a centralized model trained on the combined client and server datasets. In \method, we refer to the models uploaded by
clients as teacher models and the global model as the student model, following the knowledge distillation framework.
\subsection{Multi-client Calibration}
\label{sec:mcbnc}

\begin{algorithm}[t]
\small
\caption{Multi-client Calibration}
\label{alg:bn_calibration}
\begin{algorithmic}[1]
\REQUIRE Teacher models $\{\theta_k\}_{k=1}^{K}$,
server dataset $\mathcal{D}_s$
\ENSURE Calibrated teachers
$\{\widehat{\theta}_k\}_{k=1}^{K}$

\FOR{$k=1$ to $K$}
    \STATE Freeze trainable parameters of $\theta_k$.
    \STATE Reset BN running statistics of $\theta_k$.
    \FOR{each batch $x \in \mathcal{D}_s$}
        \STATE Forward $x$ through $\theta_k$
        without gradient.
        \STATE Update BN running mean and variance.
    \ENDFOR
    \STATE Store the calibrated teacher as
    $\widehat{\theta}_k$.
\ENDFOR

\RETURN $\{\widehat{\theta}_k\}_{k=1}^{K}$
\end{algorithmic}
\end{algorithm}

In federated learning, client models are independently trained on private datasets collected from different environments. These datasets are generally non-independent and identically distributed (Non-IID), as road conditions and driving behaviors vary across cities, leading to distinct feature distributions across clients. Consequently, the running statistics accumulated by BN layers during local training may not adapt to the distributions of unseen datasets. Directly applying multiple client models to the distillation dataset can introduce domain-dependent feature discrepancies, potentially affecting feature fusion. Once uploaded to the server, these client models serve as teacher models for subsequent knowledge distillation. 

To mitigate this issue, we propose multi-client
calibration to adapt the BN statistics of each teacher model to a shared proxy domain before feature fusion and knowledge distillation. Unlike existing methods that either preserve client-specific BN statistics \cite{li2021fedbn} or aggregate BN
statistics across clients \cite{zhong2023making},
our approach independently recalibrates each
teacher model using the server-side distillation
dataset while keeping the teacher models'
learnable parameters frozen. 

The calibration procedure is presented in \cref{alg:bn_calibration}. 
Before calibrating the BN layers of the teacher models, all learnable parameters of each teacher model are frozen. Firstly, the distillation dataset $\mathcal{D}_s$ is forwarded through the teacher models $\theta_k$ without gradient computation. Then, for BN layers of teacher model $\theta_k$, the distillation domain mean $\boldsymbol{\mu}_{k,s}$ and variance
$\boldsymbol{\sigma}_{k,s}^2$ are estimated as
\begin{equation}
\boldsymbol{\mu}_{k,s}
= \frac{1}{N}
\sum_{i=1}^{N} \sum_{h,w}
\mathbf{x}_{k,i,h,w},
\label{eq:proxy_mean}
\end{equation}
\begin{equation}
\boldsymbol{\sigma}_{k,s}^2
= \frac{1}{N}
\sum_{i=1}^{N} \sum_{h,w}
\left(
\mathbf{x}_{k,i,h,w}
- \boldsymbol{\mu}_{k,s}
\right)^2,
\label{eq:proxy_var}
\end{equation}
where $\mathbf{x}_{k,i,h,w}$ denotes the input
features at spatial location $(h,w)$ of BN
layers in teacher model $\theta_k$, and $N$
is the total number of samples in the
distillation dataset. $s$ denotes the
server distillation domain.

The original BN running statistics are then replaced with
the distillation domain estimates. The original BN running statistics are then replaced with the distillation domain estimates. The calibrated BN for teacher models is
\begin{equation}
\widehat{\operatorname{BN}}_{k}(\mathbf{x})
=
\frac{
\mathbf{x}-\boldsymbol{\mu}_{k,s}
}{
\sqrt{
\boldsymbol{\sigma}_{k,s}^{2}+\epsilon
}
},
\label{eq:calibrated_bn}
\end{equation}
The calibrated teacher models are denoted by $\{\widehat{\theta}_k\}_{k=1}^{K}$. Each calibrated teacher keeps its original
learnable parameters $\theta_k$, with only the BN running statistics updated.

Importantly, the calibration is performed independently for each teacher model, updating only the BN running statistics without backpropagation or other modifications to model weights. Each teacher model keeps its
own calibrated statistics, preserving its internal feature representation while adapting to the common distillation domain. With this approach, we produce distinct BN statistics for different teacher models while keeping
them as close as possible to the distillation dataset. It reduces feature discrepancies caused by mismatched BN statistics when models trained on different datasets process the same distillation data, thereby facilitating more effective feature fusion. Furthermore, the entire calibration procedure requires neither additional communication rounds nor access to private client data.
\subsection{Anchor-based Alignment}
\label{sec:apa}
\begin{algorithm}[t]
\small
\caption{Anchor-based Alignment}
\label{alg:anchor_alignment}
\begin{algorithmic}[1]
\REQUIRE Calibrated teacher models
$\{\widehat{\theta}_k\}_{k=1}^{K}$,
initial student $\theta_s$,
server dataset $\mathcal{D}_s$
\ENSURE Orthogonal matrices
$\{\mathbf{R}_k^*\}_{k=1}^{K}$

\STATE Freeze all model parameters.
\STATE Initialize
$\{\mathbf{C}_k \leftarrow \mathbf{0}_{d\times d}\}_{k=1}^{K}$.

\FOR{each batch $\mathbf{x}$ in $\mathcal{D}_s$}
    \STATE Extract and normalize student features:
    \STATE $\mathbf{Z}_s \leftarrow
    \operatorname{Reshape}
    (\operatorname{LN}(\mathbf{F}_s(\mathbf{x})))$

    \FOR{$k=1$ to $K$}
        \STATE Extract and normalize teacher features:
        \STATE $\mathbf{Z}_k \leftarrow
        \operatorname{Reshape}
        (\operatorname{LN}(\mathbf{F}_k(\mathbf{x})))$

        \STATE Accumulate cross-product matrix:
        \STATE $\mathbf{C}_k \leftarrow
        \mathbf{C}_k+\mathbf{Z}_k^\top\mathbf{Z}_s$
    \ENDFOR
\ENDFOR

\FOR{$k=1$ to $K$}
    \STATE $\mathbf{U}_k\boldsymbol{\Sigma}_k
    \mathbf{V}_k^\top
    \leftarrow \operatorname{SVD}(\mathbf{C}_k)$
    \STATE $\mathbf{R}_k^*
    \leftarrow \mathbf{U}_k\mathbf{V}_k^\top$
\ENDFOR

\RETURN $\{\mathbf{R}_k^*\}_{k=1}^{K}$
\end{algorithmic}
\end{algorithm}
Multi-client BN calibration reduces distribution shifts between client local data and the server-side distillation dataset by recalibrating each client model's BN statistics. However, intermediate features across client models may still reside in distinct representation spaces, since client models are trained independently on Non-IID data, even if they use the same architecture and initialization. In particular, different teachers may encode similar driving information in different representation spaces. To address this issue, we propose anchor-based alignment, which uses the student feature space as a common reference to align the intermediate features of multiple teacher models. 

Specifically, we use the initial student model $\theta_s$ as a shared anchor to align the calibrated teacher models $\{\widehat{\theta}_k\}_{k=1}^{K}$. With all models frozen, we extract their intermediate features using the server-side distillation dataset $\mathcal{D}_s$. Let $\mathbf{F}_k(\mathbf{x}), \mathbf{F}_s(\mathbf{x}) \in \mathbb{R}^{L\times d}$ denote the teacher and student features, respectively, where $L$ is the number of tokens and $d$ is the feature dimension. For each teacher, we estimate an orthogonal transformation matrix $\mathbf{R}_k\in\mathbb{R}^{d\times d}$ to map its features into the student feature space. To reduce feature magnitude differences across models, we normalize each token along the feature dimension:
\begin{equation}
\overline{\mathbf{F}}_{k,n} = \operatorname{LN}(\mathbf{F}_{k,n}), \quad
\overline{\mathbf{F}}_{s,n} = \operatorname{LN}(\mathbf{F}_{s,n}),
\label{eq:apa_norm}
\end{equation}
where $\operatorname{LN}(\cdot)$ denotes layer normalization without learnable affine parameters, and $n$ indexes the distillation samples. The normalized features of all samples are then concatenated along the token dimension:
\begin{equation}
\mathbf{X}_k = \operatorname{Concat}_{n=1}^{N}(\overline{\mathbf{F}}_{k,n}), \quad
\mathbf{X}_s = \operatorname{Concat}_{n=1}^{N}(\overline{\mathbf{F}}_{s,n}).
\label{eq:apa_concat}
\end{equation}
where $\mathbf{X}_k,\mathbf{X}_s\in\mathbb{R}^{NL\times d}$ represent the teacher and student feature matrix, respectively. For each teacher, the optimal alignment matrix $\mathbf{R}_k^{*}$ is obtained by solving the orthogonal Procrustes problem:
\begin{equation}
\min\limits_{\mathbf{R}_k}
\|\mathbf{X}_k\mathbf{R}_k-\mathbf{X}_s\|_F^2,
\quad
\mathrm{s.t.}\;
\mathbf{R}_k^\top\mathbf{R}_k=\mathbf{I}.
\label{eq:apa_opt}
\end{equation}
where \(\|\cdot\|_F\) denotes the Frobenius norm, which measures the difference between the aligned teacher features and the student features. This optimization finds an orthogonal transformation that minimizes the difference between the teacher features and their corresponding student features. The problem admits a closed-form solution using singular value decomposition (SVD) \cite{schonemann1966generalized}:
\begin{equation}
\mathbf{X}_k^\top\mathbf{X}_s = \mathbf{U}_k\boldsymbol{\Sigma}_k \mathbf{V}_k^\top,
\label{eq:apa_svd}
\end{equation}
where \(\mathbf U_k\) and \(\mathbf V_k\) contain the left and right singular vectors, respectively, and \(\boldsymbol\Sigma_k\) is the diagonal matrix of singular values. The optimal transformation matrix is given by
$\mathbf{R}_k^{*} = \mathbf{U}_k\mathbf{V}_k^\top$. The complete alignment procedure is shown in \cref{alg:anchor_alignment}. Since $\mathbf{R}_k^*$ is orthogonal, it preserves the pairwise Euclidean distances between features within each teacher \cite{gower2004procrustes}. For any two normalized token features $\overline{\mathbf{f}}_{k,i}$ and $\overline{\mathbf{f}}_{k,j}$, the Euclidean distance is preserved under the orthogonal transformation:
\begin{equation}
\left\|
\overline{\mathbf{f}}_{k,i}\mathbf{R}_k^*
-
\overline{\mathbf{f}}_{k,j}\mathbf{R}_k^*
\right\|_2
=
\left\|
\overline{\mathbf{f}}_{k,i}
-
\overline{\mathbf{f}}_{k,j}
\right\|_2.
\label{eq:apa_distance}
\end{equation}
Thus, our approach maps teacher features into the student representation space while preserving their geometric structure. The transformation matrices are computed offline and remain fixed during knowledge distillation. For each distillation sample, the aligned teacher features are obtained as
\begin{equation}
\widetilde{\mathbf{F}}_{k,n}
= \operatorname{LN}(\mathbf{F}_{k,n})\mathbf{R}_k^*.
\end{equation}
Compared with pairwise teacher alignment \cite{achara2026multiway}, which requires \(\mathcal{O}(K^2)\) mappings, our approach requires only \(K\) transformation matrices. By aligning teacher features with the initial student representation space, this approach not only facilitates subsequent feature comparison and fusion but also introduces no additional learnable parameters, while keeping all teacher models frozen during distillation. The aligned features are then used for token-wise consensus distillation.
\subsection{Token-wise Consensus Distillation}
\label{sec:tcd}
\begin{figure*}[t]
    \centering
    \includegraphics[width=1\textwidth]{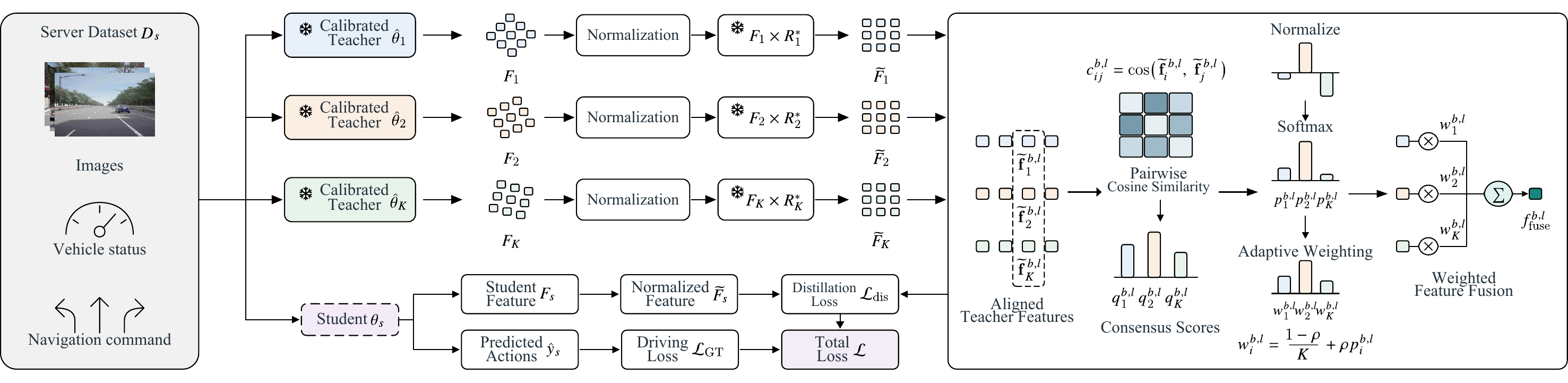}
    \caption{Overview of token-wise consensus distillation. The server dataset is first fed into the calibrated teacher models and the student model to extract features. The teacher features are then mapped into the student feature space using the optimal orthogonal transformation matrices. Next, token-wise consensus is used to compute adaptive weights for teacher feature fusion. Finally, the student model is trained by minimizing the distillation loss between the fused teacher features and student features, together with the supervised driving action loss.}
    \label{fig:tcd}
\end{figure*}
Following multi-client calibration and anchor-based alignment, the teacher features are adapted to the server-side distillation domain and aligned with the student representation space. However, teachers trained independently on Non-IID driving data may still provide varying levels of useful information across different features. Simply averaging their features assigns equal importance to all teachers and fails to account for these differences. 

To address this issue, we propose token-wise consensus distillation, which assigns adaptive weights to teacher features based on their agreement across teachers at each token, as illustrated in \cref{fig:tcd}. The underlying assumption is that teacher features with higher agreement are more likely to capture knowledge shared across clients. First, we extract features from the calibrated teacher models $\{\widehat{\theta}_k\}_{k=1}^{K}$ using the server-side distillation data and align them with the student feature space using the fixed orthogonal matrices obtained from anchor-based alignment. 
The aligned teacher features
$\widetilde{\mathbf{F}}_k \in \mathbb{R}^{B\times L\times d}$
are $L2$ normalized at each token to reduce scale differences,
where $B$, $L$, and $d$ denote the batch size, number of tokens,
and feature dimension, respectively:
\begin{equation}
\mathbf{z}_k^{b,l} =
\frac{\widetilde{\mathbf{f}}_k^{b,l}}
{\|\widetilde{\mathbf{f}}_k^{b,l}\|_2+\epsilon},
\label{eq:tcd_l2}
\end{equation}
where $\epsilon$ is a small constant for numerical stability. The normalized features are then used to compute the pairwise cosine similarity between teachers $i$ and $j$ at each token:
\begin{equation}
c_{i,j}^{b,l}= (\mathbf{z}_i^{b,l})^\top \mathbf{z}_j^{b,l}, \quad i \neq j.
\label{eq:tcd_similarity}
\end{equation}
For each token in a given sample, we compute the consensus score of each teacher as its average similarity with all other teachers:
\begin{equation}
q_i^{b,l} = \frac{1}{K-1} \sum_{\substack{j=1\\j\neq i}}^{K} c_{i,j}^{b,l}.
\label{eq:tcd_consensus}
\end{equation}
A higher consensus score indicates greater agreement between a teacher's features and those of other teachers at the corresponding token. To make the consensus scores comparable across teachers, we normalize them separately for each token:
\begin{equation}
\widehat{q}_i^{b,l}
=
\frac{q_i^{b,l}-\mu_q^{b,l}}
{\sigma_q^{b,l}+\epsilon},
\label{eq:tcd_score_norm}
\end{equation}
where $\mu_q^{b,l}$ and $\sigma_q^{b,l}$ denote the mean and standard deviation of consensus scores across teachers for the same sample and token,
respectively, and $\epsilon$ is a small constant for numerical stability. The normalized consensus scores are then converted into teacher weights using Softmax:
\begin{equation}
p_i^{b,l}
=
\frac{\exp(\widehat{q}_i^{b,l}/\tau)}
{\sum_{j=1}^{K}\exp(\widehat{q}_j^{b,l}/\tau)},
\label{eq:tcd_softmax}
\end{equation}
where $\tau$ is a temperature parameter controlling the sharpness of the weight distribution. However, relying exclusively on consensus-based weights may underestimate the contributions of teachers with lower agreement, potentially limiting the diversity of knowledge. To balance consensus-based weights and uniform teacher contributions, we introduce an adaptive weighting strategy:
\begin{equation}
w_i^{b,l} = \frac{1-\rho}{K}+\rho p_i^{b,l}.
\label{eq:tcd_adaptive}
\end{equation}
where $\rho\in[0,1]$ controls the contribution of consensus-based weighting. Since the Softmax weights satisfy $\sum_{i=1}^{K}p_i^{b,l}=1$, the adaptive weights remain normalized:
\begin{equation}
\sum_{i=1}^{K}w_i^{b,l}
= (1-\rho)+\rho\sum_{i=1}^{K}p_i^{b,l}=1.
\label{eq:tcd_weight_sum}
\end{equation}

This strategy preserves a minimum contribution from each teacher while assigning higher weights to teachers with greater feature agreement. 
The aligned teacher features are fused using the adaptive weights
to construct the student distillation target as
$\mathbf{f}_{\mathrm{fuse}}^{b,l} =
\sum_{i=1}^{K} w_i^{b,l}\widetilde{\mathbf{f}}_i^{b,l}$. Since the orthogonal transformation matrices are estimated using LayerNorm for student features, we apply the same parameter-free LayerNorm to the student features during distillation, yielding $\widetilde{\mathbf{f}}_s^{b,l} = \operatorname{LN}(\mathbf{f}_s^{b,l})$. We use the L1 distance between the normalized student features and the fused teacher features as the distillation loss:
\begin{equation}
\mathcal{L}_{\mathrm{dis}} = \frac{1}{BLd} \sum_{b=1}^{B} \sum_{l=1}^{L} \left\|
\widetilde{\mathbf{f}}_s^{b,l} - \mathbf{f}_{\mathrm{fuse}}^{b,l} \right\|_1.
\label{eq:tcd_loss}
\end{equation}
During distillation, all teacher models and their alignment matrices are frozen, while only the student model is updated. Besides distillation loss, we incorporate supervised learning loss using ground-truth driving actions
from the server distillation dataset. The total loss
is defined as
$\mathcal{L} = \alpha\mathcal{L}_{\mathrm{GT}} (\widehat{\mathbf{y}}_s,\mathbf{y}_s) + \gamma\mathcal{L}_{\mathrm{dis}}$, where $\alpha$ and $\gamma$ determine the relative contributions of the supervised and distillation losses. 

\section{Experiments}
\label{sec:experiments}
\subsection{Datasets}
We establish a Non-IID federated learning setting using data from four towns in the Bench2Drive benchmark \cite{jia2024bench}: Town03, Town12, Town13, and Town15, containing 81,276, 94,111, 125,655, and 72,459 training samples, respectively. Each sample consists of synchronized front and rear RGB images, vehicle speed, a high-level navigation command, and the corresponding steering, throttle, and brake inputs. The original RGB images are resized from 1600 × 900 to 400 × 225 pixels before being fed into the model. In terms of route and scenario coverage, the four towns contain 272, 429, 695, and 318 routes, spanning 28, 42, 41, and 32 scenario categories, respectively.
\subsection{Implementation Details}
\noindent\textbf{Client model.} All client models employed CIL++~\cite{xiao2023scaling}, a pure vision-based end-to-end driving architecture with a ResNet34 backbone~\cite{he2016deep} pretrained on ImageNet-1K~\cite{russakovsky2015imagenet}. The model takes RGB images, normalized vehicle speed and a one-hot navigation command as inputs to predict steering and acceleration, with acceleration defined as throttle minus brake. Each client model is independently trained for 80 epochs with a batch size of 256 using AdamW \cite{loshchilov2019decoupled}, an initial learning rate of $1\times10^{-4}$, and weight decay of 0.01. The learning rate is halved at epochs 30, 50, and 65, with a minimum of $1\times10^{-5}$. 

\noindent\textbf{Global model.} The global model is trained through knowledge distillation using the same architecture and training settings as client models, except for an initial learning rate of $1\times10^{-5}$ and a minimum learning rate of $1\times10^{-6}$. Both Anchor-based alignment and token-wise consensus operate on features from the final Transformer~\cite{vaswani2017attention} encoder. The consensus temperature and adaptive weighting parameter are set to $\tau=1$ and $\rho=0.5$, respectively. The weights of the supervised driving loss $\alpha$ and feature distillation loss $\gamma$ are both set to 1.

\noindent\textbf{Federated settings.} For \method and FedDF, one dataset is used for server-side distillation, while the remaining three are assigned to three separate clients. For the other reproduced federated learning methods, each of the four datasets is assigned to a separate client.

\noindent\textbf{Evaluation Metrics.} We evaluate all models using CARLA 0.9.15 and the Bench2Drive benchmark, comprising 220 evaluation routes. The primary metrics include Driving Score (DS), Success Rate (SR), Route Completion (RC), and Infraction Penalty (IP). 
DS combines route completion and infraction penalties; SR measures the percentage of routes successfully completed; RC represents the average route completion percentage; and IP measures the cumulative penalty factor associated with driving infractions, with higher values indicating fewer penalties. 
We additionally report the macro-average performance across five driving abilities: Overtaking, Merging, Emergency Brake, Give Way, and Traffic Signs.

\subsection{One-Shot Federated Learning}
\subsubsection{Main Metrcis}
\label{sec:one_shot}
\textbf{Comparison with Federated Baselines.}
\begin{table}[t]
\centering
\caption{One-shot federated learning results on Bench2Drive over three random seeds. For \method and FedDF, Town 12 is used for server-side distillation, while Town 03, Town 13 and Town 15 are assigned to three separate clients. For the other reproduced federated learning methods, all four datasets are assigned to four separate clients. Centralized training on all four towns serves as the performance upper bound.}
\label{tab:oneshot}
\setlength{\tabcolsep}{12pt}
\renewcommand{\arraystretch}{1.15}
\resizebox{\columnwidth}{!}{%
\begin{tabular}{lcccc}
\toprule
Method & DS $\uparrow$ & SR (\%) $\uparrow$ & RC (\%) $\uparrow$ & IP $\uparrow$ \\
\midrule
FedAvg \cite{mcmahan2017communication}
& 7.05 $\pm$ 0.12
& 0.00 $\pm$ 0.00
& 8.45 $\pm$ 0.16
& \textbf{0.93 $\pm$ 0.00} \\

FedRDN \cite{yan2025simple}
& 14.99 $\pm$ 0.09
& 0.00 $\pm$ 0.00
& 44.75 $\pm$ 0.28
& 0.36 $\pm$ 0.00 \\

FedProx \cite{li2020federated}
& 26.32 $\pm$ 0.35
& 1.82 $\pm$ 0.00
& 58.39 $\pm$ 0.59
& 0.47 $\pm$ 0.00 \\

FedDF \cite{lin2020ensemble}
& 51.78 $\pm$ 0.50
& 20.61 $\pm$ 0.57
& 69.73 $\pm$ 0.06
& 0.73 $\pm$ 0.01 \\

FedE2E (Ours)
& \textbf{55.85 $\pm$ 0.23}
& \textbf{24.70 $\pm$ 0.93}
& \textbf{71.51 $\pm$ 0.56}
& 0.78 $\pm$ 0.01 \\
\midrule
Centralized
& 63.30 $\pm$ 0.63
& 29.24 $\pm$ 0.57
& 80.08 $\pm$ 0.69
& 0.78 $\pm$ 0.01 \\
\bottomrule
\end{tabular}%
}
\end{table}
\cref{tab:oneshot} presents the closed-loop driving performance of \method with conventional federated learning methods after one communication round. FedE2E achieves a DS of 55.85, outperforming all reproduced federated baselines. Compared with FedDF, which also employs knowledge distillation, our method improves DS by 4.07 points, SR by 4.09 percentage points, and RC by 1.78 percentage points. It also achieves an IP of 0.78, matching the centralized reference. Furthermore, FedE2E achieves a DS standard deviation of 0.23, lower than 0.50 for FedDF and 0.63 for the centralized reference, suggesting more consistent performance. The lower scores of FedAvg, FedRDN, and FedProx highlight the limitations of averaging parameter aggregation under heterogeneous client distributions in the one-shot setting.

\begin{table}[t]
\setlength{\tabcolsep}{12pt}
\centering
\caption{Main metrics of client models.}
\label{tab:local}
\renewcommand{\arraystretch}{1.15}
\resizebox{\columnwidth}{!}{%
\begin{tabular}{lcccc}
\toprule
Model & DS $\uparrow$ & SR (\%) $\uparrow$ & RC (\%) $\uparrow$ & IP $\uparrow$ \\
\midrule
Town03
& 41.88 $\pm$ 0.64
& 14.55 $\pm$ 1.62
& 58.33 $\pm$ 0.63
& 0.70 $\pm$ 0.00 \\

Town12
& 46.89 $\pm$ 0.62
& 14.24 $\pm$ 0.93
& 69.04 $\pm$ 0.88
& 0.68 $\pm$ 0.00 \\

Town13
& 50.01 $\pm$ 0.72
& 18.79 $\pm$ 1.55
& 69.52 $\pm$ 0.61
& 0.71 $\pm$ 0.00 \\

Town15
& 29.84 $\pm$ 0.97
& 4.24 $\pm$ 0.21
& 52.12 $\pm$ 0.97
& 0.56 $\pm$ 0.01 \\

FedE2E (Ours)
& \textbf{55.85 $\pm$ 0.23}
& \textbf{24.70 $\pm$ 0.93}
& \textbf{71.51 $\pm$ 0.56}
& \textbf{0.78 $\pm$ 0.01} \\
\midrule
Centralized
& 63.30 $\pm$ 0.63
& 29.24 $\pm$ 0.57
& 80.08 $\pm$ 0.69
& 0.78 $\pm$ 0.01 \\
\bottomrule
\end{tabular}%
}
\end{table}
\noindent\textbf{Comparison with Client Models.} \cref{tab:local} compares FedE2E with independently trained client models. FedE2E outperforms all four client models across all evaluation metrics. Compared with the best client model, Town13, FedE2E improves DS by 5.84 points and SR by 5.91 points, demonstrating the benefit of knowledge transfer across multiple client models. Although FedE2E remains 7.45 DS points below the centralized reference, it achieves these results without accessing private client datasets and requires only a single communication round.
\subsubsection{Driving Ability}
\label{sec:driving_ability}

\begin{table}[t]
\centering
\caption{Driving abilities of different federated methods.}
\label{tab:driving_ability}
\renewcommand{\arraystretch}{1.15}
\resizebox{\columnwidth}{!}{%
\begin{tabular}{lcccccc}
\toprule
Method
& Overtaking $\uparrow$
& Merging $\uparrow$
& \headercell{Emergency\\Brake}
& \headercell{Give\\Way}
& \headercell{Traffic\\Signs}
& \headercell{Ability\\Avg.} \\
\midrule

FedAvg
& 0.00 $\pm$ 0.00
& 0.00 $\pm$ 0.00
& 0.00 $\pm$ 0.00
& 0.00 $\pm$ 0.00
& 4.91 $\pm$ 0.25
& 0.98 $\pm$ 0.05 \\

FedProx
& 0.00 $\pm$ 0.00
& 2.50 $\pm$ 0.00
& 3.33 $\pm$ 0.00
& 0.00 $\pm$ 0.00
& 15.09 $\pm$ 0.25
& 4.18 $\pm$ 0.05 \\

FedRDN
& 0.00 $\pm$ 0.00
& 0.00 $\pm$ 0.00
& 0.00 $\pm$ 0.00
& 0.00 $\pm$ 0.00
& \textbf{30.35 $\pm$ 0.66}
& 6.07 $\pm$ 0.13 \\

FedDF
& 18.52 $\pm$ 4.19
& 19.58 $\pm$ 1.18
& 23.89 $\pm$ 0.79
& \textbf{56.67 $\pm$ 9.43}
& 17.19 $\pm$ 1.79
& 27.17 $\pm$ 1.24 \\

FedE2E (Ours)
& \textbf{28.15 $\pm$ 1.05}
& \textbf{23.75 $\pm$ 1.77}
& \textbf{28.33 $\pm$ 2.36}
& 46.67 $\pm$ 17.00
& 24.04 $\pm$ 1.38
& \textbf{30.19 $\pm$ 3.10} \\
\midrule
Centralized
& 33.33 $\pm$ 1.81
& 28.75 $\pm$ 0.00
& 33.33 $\pm$ 2.36
& 36.67 $\pm$ 4.71
& 27.19 $\pm$ 1.31
& 31.86 $\pm$ 1.28 \\
\bottomrule
\end{tabular}%
}
\end{table}
\textbf{Comparison with Federated Baselines.} To further evaluate driving performance, we analyze five driving abilities: Overtaking, Merging, Emergency Brake, Give Way, and Traffic Signs. As shown in \cref{tab:driving_ability}, FedE2E achieves the best results among federated methods in Overtaking, Merging, and Emergency Brake, outperforming FedDF by 9.63, 4.17, and 4.44 points, respectively. Although FedDF performs better in Give Way and FedRDN achieves the highest score in Traffic Signs, FedE2E obtains the highest ability macro-average of 30.19\%, exceeding FedDF by 3.02 points and approaching the centralized reference of 31.86\%. These results indicate that the improvements extend beyond route completion to multiple driving abilities.

\begin{table}[t]
\centering
\caption{Driving abilities of client models.}
\label{tab:local_ability}
\renewcommand{\arraystretch}{1.15}
\resizebox{\columnwidth}{!}{%
\begin{tabular}{lcccccc}
\toprule
Method
& Overtaking $\uparrow$
& Merging $\uparrow$
& \headercell{Emergency\\Brake}
& \headercell{Give\\Way}
& \headercell{Traffic\\Signs}
& \headercell{Ability\\Avg.} \\
\midrule
Town03
& 9.63 $\pm$ 1.05
& 12.08 $\pm$ 1.18
& 18.89 $\pm$ 4.16
& \textbf{63.33 $\pm$ 4.71}
& 16.14 $\pm$ 1.08
& 24.02 $\pm$ 1.78 \\

Town12
& 17.04 $\pm$ 5.24
& 14.58 $\pm$ 1.56
& 13.33 $\pm$ 1.36
& 30.00 $\pm$ 8.16
& 20.00 $\pm$ 2.27
& 18.99 $\pm$ 2.50 \\

Town13
& 15.56 $\pm$ 4.80
& 19.17 $\pm$ 0.59
& 21.11 $\pm$ 1.57
& 43.33 $\pm$ 4.71
& 17.19 $\pm$ 1.79
& 23.27 $\pm$ 1.40 \\

Town15
& 1.48 $\pm$ 1.05
& 2.50 $\pm$ 0.00
& 6.11 $\pm$ 0.79
& 16.67 $\pm$ 4.71
& 11.75 $\pm$ 0.66
& 7.70 $\pm$ 0.95 \\

FedE2E (Ours)
& \textbf{28.15 $\pm$ 1.05}
& \textbf{23.75 $\pm$ 1.77}
& \textbf{28.33 $\pm$ 2.36}
& 46.67 $\pm$ 17.00
& \textbf{24.04 $\pm$ 1.38}
& \textbf{30.19 $\pm$ 3.10} \\
\midrule
Centralized
& 33.33 $\pm$ 1.81
& 28.75 $\pm$ 0.00
& 33.33 $\pm$ 2.36
& 36.67 $\pm$ 4.71
& 27.19 $\pm$ 1.31
& 31.86 $\pm$ 1.28 \\
\bottomrule
\end{tabular}%
}
\end{table}
\noindent\textbf{Comparison with Client Models.} \cref{tab:local_ability} presents FedE2E with client models across five driving abilities. The proposed method achieves the highest ability macro-average of 30.19\%, outperforming the best client model, Town03, by 6.17 points. It also achieves the best performance in four of the five driving abilities, with Town03 performing better in the Give Way ability. These results demonstrate that distilling knowledge from multiple client models can improve overall driving capabilities.

\subsection{Multi-Round Federated Learning}
\label{sec:multi_round}
\begin{table}[t]
\setlength{\tabcolsep}{12pt}
\centering
\caption{Main metrics of federated methods with five rounds.}
\label{tab:multiround}
\renewcommand{\arraystretch}{1.15}
\resizebox{\columnwidth}{!}{%
\begin{tabular}{lcccc}
\toprule
Method & DS $\uparrow$ & SR (\%) $\uparrow$ & RC (\%) $\uparrow$ & IP $\uparrow$ \\
\midrule
FedAvg
& 58.54 $\pm$ 1.40
& 25.00 $\pm$ 1.34
& 78.20 $\pm$ 1.14
& 0.74 $\pm$ 0.01 \\

FedProx
& 26.86 $\pm$ 0.64
& 2.58 $\pm$ 0.57
& 59.65 $\pm$ 0.36
& 0.46 $\pm$ 0.01 \\

FedRDN
& 45.94 $\pm$ 0.80
& 14.24 $\pm$ 1.30
& 72.13 $\pm$ 0.87
& 0.63 $\pm$ 0.00 \\

FedDF
& 58.77 $\pm$ 0.56
& 26.67 $\pm$ 0.21
& 74.79 $\pm$ 0.55
& \textbf{0.77 $\pm$ 0.01} \\

FedE2E (ours)
& \textbf{60.35 $\pm$ 0.62}
& \textbf{26.97 $\pm$ 0.77}
& \textbf{79.52 $\pm$ 0.66}
& 0.75 $\pm$ 0.01 \\
\bottomrule
\end{tabular}%
}
\end{table}
Although FedE2E is designed for one-shot federated learning, the trained global model can also serve as an initialization for subsequent federated training. To evaluate this setting, we apply FedAvg after the initial FedE2E fusion, as it is a widely adopted aggregation method in federated learning~\cite{mcmahan2017communication,li2020federated,yan2025simple,zhang2021endtoend,aparna2021steering,zhang2021realtime}. As shown in \cref{tab:multiround}, after five communication rounds, our method achieves a Driving Score of 60.35, exceeding FedAvg and FedDF by 1.81 and 1.58 points, respectively. It also achieves the highest Route Completion of 79.52\% among the compared methods. These results suggest that the global model obtained through one-shot knowledge fusion provides an effective initialization for subsequent federated optimization. 

\begin{table}[t]
\setlength{\tabcolsep}{12pt}
\centering
\caption{Main metrics of \method at each round}
\label{tab:round_progress}
\renewcommand{\arraystretch}{1.15}
\resizebox{\columnwidth}{!}{%
\begin{tabular}{ccccc}
\toprule
Round & DS $\uparrow$ & SR (\%) $\uparrow$ & RC (\%) $\uparrow$ & IP $\uparrow$ \\
\midrule
1
& 55.85 $\pm$ 0.23
& 24.70 $\pm$ 0.93
& 71.51 $\pm$ 0.56
& \textbf{0.78 $\pm$ 0.01} \\

2
& 56.84 $\pm$ 0.32
& 21.21 $\pm$ 1.50
& 77.22 $\pm$ 0.05
& 0.74 $\pm$ 0.01 \\

3
& 59.58 $\pm$ 0.38
& 25.61 $\pm$ 1.67
& 78.96 $\pm$ 0.62
& 0.75 $\pm$ 0.00 \\

4
& 58.32 $\pm$ 0.80
& 25.15 $\pm$ 1.83
& 78.09 $\pm$ 0.17
& 0.74 $\pm$ 0.01 \\

5
& \textbf{60.35 $\pm$ 0.62}
& \textbf{26.97 $\pm$ 0.77}
& \textbf{79.52 $\pm$ 0.66}
& 0.75 $\pm$ 0.01 \\
\bottomrule
\end{tabular}%
}
\end{table}
\cref{tab:round_progress} further presents the performance of \method across different communication rounds. Notably, it achieves a Driving Score of 59.58 after only three communication rounds, already exceeding the best five-round result among the baselines, achieved by FedDF at 58.77. This reduces the required communication rounds from five to three, corresponding to a 40\% reduction in the number of rounds required to reach this performance level.

\subsection{Impact of Distillation Datasets}
\label{sec:distillation_data}
\begin{table}[t]
\setlength{\tabcolsep}{8pt}
\centering
\caption{Comparison of FedE2E and FedDF using different towns as 
server distillation datasets.}
\label{tab:distillation_dataset}
\renewcommand{\arraystretch}{1.15}
\resizebox{\columnwidth}{!}{%
\begin{tabular}{llcccc}
\toprule
Distillation & Method & DS $\uparrow$ & SR (\%) $\uparrow$ & RC (\%) $\uparrow$ & IP $\uparrow$ \\
\midrule
\multirow{2}{*}{Town03}
& FedE2E
& \textbf{57.47 $\pm$ 0.71}
& \textbf{24.09 $\pm$ 1.48}
& \textbf{71.31 $\pm$ 0.54}
& \textbf{0.80 $\pm$ 0.00} \\
& FedDF
& 49.42 $\pm$ 0.68
& 19.09 $\pm$ 0.37
& 65.77 $\pm$ 0.06
& 0.73 $\pm$ 0.01 \\
\midrule
\multirow{2}{*}{Town12}
& FedE2E
& \textbf{55.85 $\pm$ 0.23}
& \textbf{24.70 $\pm$ 0.93}
& \textbf{71.51 $\pm$ 0.56}
& \textbf{0.78 $\pm$ 0.01} \\
& FedDF
& 51.78 $\pm$ 0.50
& 20.61 $\pm$ 0.57
& 69.73 $\pm$ 0.06
& 0.73 $\pm$ 0.01 \\
\midrule
\multirow{2}{*}{Town13}
& FedE2E
& \textbf{54.12 $\pm$ 0.85}
& \textbf{20.15 $\pm$ 1.13}
& 66.76 $\pm$ 0.58
& \textbf{0.82 $\pm$ 0.01} \\
& FedDF
& 49.47 $\pm$ 0.40
& 17.12 $\pm$ 0.43
& \textbf{68.58 $\pm$ 0.70}
& 0.70 $\pm$ 0.00 \\
\midrule
\multirow{2}{*}{Town15}
& FedE2E
& \textbf{54.03 $\pm$ 0.49}
& 21.36 $\pm$ 0.37
& \textbf{70.70 $\pm$ 0.42}
& \textbf{0.76 $\pm$ 0.01} \\
& FedDF
& 51.40 $\pm$ 0.11
& \textbf{22.73 $\pm$ 0.98}
& 67.65 $\pm$ 0.43
& 0.74 $\pm$ 0.00 \\
\bottomrule
\end{tabular}%
}
\end{table}
To assess the impact of the server-side distillation dataset, we conduct experiments with different server-side distillation datasets, while using the remaining datasets for client training. As shown in \cref{tab:distillation_dataset}, FedE2E consistently outperforms FedDF in DS across all four settings, with improvements ranging from 2.63 to 8.05 points. The highest DS, 57.47, is achieved with Town03, exceeding FedDF by 8.05 points. FedE2E also achieves higher SR and IP across three of the four settings, and outperforms FedDF on RC in all settings except Town13. Overall, although the server-side distillation dataset affects the performance of FedE2E, our method consistently outperforms the reproduced federated learning methods reported in \cref{tab:oneshot}.

\subsection{Ablation Study}
\label{sec:ablation}
\begin{table}[t]
\setlength{\tabcolsep}{8pt}
\centering
\caption{Ablation study of the proposed FedE2E framework. MCC, ABA, and TCD denote multi-client calibration, anchor-based alignment, and token-wise consensus distillation, respectively.}
\label{tab:ablation}
\renewcommand{\arraystretch}{1.15}
\resizebox{\columnwidth}{!}{%
\begin{tabular}{ccccccc}
\toprule
MCC & ABA & TCD &
DS $\uparrow$ & SR (\%) $\uparrow$ & RC (\%) $\uparrow$ & IP $\uparrow$ \\
\midrule
-- & \checkmark & \checkmark
& 52.82 $\pm$ 0.48
& 20.45 $\pm$ 0.74
& 70.49 $\pm$ 0.61
& 0.75 $\pm$ 0.00 \\

\checkmark & -- & \checkmark
& 50.80 $\pm$ 0.75
& 17.88 $\pm$ 2.39
& 69.22 $\pm$ 0.27
& 0.75 $\pm$ 0.01 \\

\checkmark & \checkmark & --
& 54.46 $\pm$ 1.25
& 23.48 $\pm$ 1.19
& 70.58 $\pm$ 0.77
& 0.77 $\pm$ 0.01 \\

\checkmark & \checkmark & \checkmark
& \textbf{55.85 $\pm$ 0.23}
& \textbf{24.70 $\pm$ 0.93}
& \textbf{71.51 $\pm$ 0.56}
& \textbf{0.78 $\pm$ 0.01} \\
\bottomrule
\end{tabular}%
}
\end{table}
\cref{tab:ablation} presents the contribution of each component of \method. Multi-client calibration improves DS by 3.03 points by recalibrating the BN statistics of client models using server-side distillation data, helping reduce statistical differences between client and server domains. Anchor-based alignment provides the largest performance gain, improving DS by 5.05 points and SR by 6.82 percentage points. Mapping independently trained teacher features into a shared student-anchored space makes their representations more comparable before knowledge fusion. Token-wise consensus distillation further improves DS by 1.39 points and reduces its standard deviation from 1.25 to 0.23. By adaptively weighting teacher features based on their token-wise consistency, this approach improves knowledge fusion while preserving the contributions of teachers with lower consensus scores. Overall, the three components address different challenges in multi-client knowledge distillation, and their combination achieves the best performance across all evaluation metrics.
\section{Conclusion}
\label{conclusion}
This paper proposes \method, the first one-shot federated learning framework tailored to end-to-end autonomous driving. \method integrates multi-client calibration, anchor-based feature alignment, and token-wise consensus distillation to effectively distill heterogeneous knowledge from multiple client models without accessing their private data. Extensive experiments and closed-loop evaluation on Bench2Drive demonstrate that the global model outperforms individual client models and prior federated learning methods in both one-shot and multi-round settings. Future work will explore scaling \method to more powerful end-to-end driving architectures and evaluating its generalization in more diverse driving environments.
{
    \small
    \bibliographystyle{IEEEtran}
    \bibliography{main}

@String(CVPR= {IEEE Conf. Comput. Vis. Pattern Recog.})

@String(ICCV= {Int. Conf. Comput. Vis.})

@String(AAAI = {AAAI})

@String(CVPR  = {CVPR})

@String(ICCV  = {ICCV})

@article{chellapandi2024federated,
  title     = {Federated Learning for Connected and Automated
               Vehicles: A Survey of Existing Approaches and Challenges},
  author    = {Chellapandi, Vishnu Pandi
               and Yuan, Liangqi
               and Brinton, Christopher G.
               and Zak, Stanislaw H.
               and Wang, Ziran},
  journal   = {IEEE Transactions on Intelligent Vehicles},
  volume    = {9},
  number    = {1},
  pages     = {119--137},
  year      = {2024},
  doi       = {10.1109/TIV.2023.3332675}
}

@article{kalloori2024crosssilo,
  title     = {Towards Cross-Silo Federated Learning
               for Corporate Organizations},
  author    = {Kalloori, Saikishore
               and Srivastava, Abhishek},
  journal   = {Knowledge-Based Systems},
  volume    = {289},
  pages     = {111501},
  year      = {2024},
  doi       = {10.1016/j.knosys.2024.111501}
}

@inproceedings{zhang2022dense,
  title     = {{DENSE}: Data-Free One-Shot Federated Learning},
  author    = {Zhang, Jie and Chen, Chen and Li, Bo
               and Lyu, Lingjuan and Wu, Shuang
               and Ding, Shouhong and Shen, Chunhua
               and Wu, Chao},
  booktitle = {Advances in Neural Information Processing Systems},
  volume    = {35},
  pages     = {21414--21428},
  year      = {2022}
}

@inproceedings{lian2026feds2r,
  title     = {{FedS2R}: One-Shot Federated Domain
               Generalization for Synthetic-to-Real
               Semantic Segmentation in Autonomous Driving},
  author    = {Lian, Tao and G{\'o}mez, Jose Luis
               and L{\'o}pez, Antonio M.},
  booktitle = {2026 IEEE Intelligent Vehicles Symposium (IV)},
  pages     = {1061--1068},
  year      = {2026},
  publisher = {IEEE},
  doi       = {10.1109/IV66570.2026.11624094}
}

@inproceedings{yan2025simple,
  title={A Simple Data Augmentation for Feature Distribution
         Skewed Federated Learning},
  author={Yan, Yunlu and Fu, Huazhu and Li, Yuexiang
          and Xie, Jinheng and Ma, Jun and Yang, Guang
          and Zhu, Lei},
  booktitle={Proceedings of the IEEE/CVF Conference on
             Computer Vision and Pattern Recognition (CVPR)},
  pages={25749--25758},
  year={2025},
  doi={10.1109/CVPR52734.2025.02398}
}

@inproceedings{yang2024feddeo,
  title     = {{FedDEO}: Description-Enhanced One-Shot
               Federated Learning with Diffusion Models},
  author    = {Yang, Mingzhao
               and Su, Shangchao
               and Li, Bin
               and Xue, Xiangyang},
  booktitle = {Proceedings of the 32nd ACM International
               Conference on Multimedia},
  pages     = {6666--6675},
  year      = {2024},
  doi       = {10.1145/3664647.3681490}
}

@inproceedings{heinbaugh2023datafree,
  title     = {Data-Free One-Shot Federated Learning
               Under Very High Statistical Heterogeneity},
  author    = {Heinbaugh, Clare Elizabeth
               and Luz-Ricca, Emilio
               and Shao, Huajie},
  booktitle = {International Conference on Learning Representations},
  year      = {2023},
  url       = {https://openreview.net/forum?id=hHBWwN8f7u}
}

@inproceedings{sun2023feature,
  title     = {Feature Distribution Matching for Federated Domain Generalization},
  author    = {Sun, Yuwei and Chong, Ng and Ochiai, Hideya},
  booktitle = {Proceedings of the 14th Asian Conference on Machine Learning},
  series    = {Proceedings of Machine Learning Research},
  volume    = {189},
  pages     = {942--957},
  year      = {2023},
  publisher = {PMLR},
  url       = {https://proceedings.mlr.press/v189/sun23a.html}
}

@inproceedings{mcmahan2017communication,
  title     = {Communication-Efficient Learning of Deep Networks
               from Decentralized Data},
  author    = {McMahan, Brendan and Moore, Eider
               and Ramage, Daniel and Hampson, Seth
               and Arcas, Blaise Aguera y},
  booktitle = {Proceedings of the 20th International
               Conference on Artificial Intelligence
               and Statistics (AISTATS)},
  series    = {Proceedings of Machine Learning Research},
  volume    = {54},
  pages     = {1273--1282},
  year      = {2017},
  publisher = {PMLR}
}

@inproceedings{zhang2023federated,
  title     = {Federated Domain Generalization
               With Generalization Adjustment},
  author    = {Zhang, Ruipeng and Xu, Qinwei
               and Yao, Jiangchao and Zhang, Ya
               and Tian, Qi and Wang, Yanfeng},
  booktitle = {Proceedings of the IEEE/CVF Conference
               on Computer Vision and Pattern Recognition (CVPR)},
  pages     = {3954--3963},
  year      = {2023},
  doi       = {10.1109/CVPR52729.2023.00385}
}

@inproceedings{chen2023federated,
  title     = {Federated Domain Generalization for Image
               Recognition via Cross-Client Style Transfer},
  author    = {Chen, Junming and Jiang, Meirui
               and Dou, Qi and Chen, Qifeng},
  booktitle = {Proceedings of the IEEE/CVF Winter Conference
               on Applications of Computer Vision (WACV)},
  pages     = {361--370},
  year      = {2023},
  doi       = {10.1109/WACV56688.2023.00044}
}

@article{yu2024contrastive,
  title     = {Contrastive-Enhanced Domain Generalization
               With Federated Learning},
  author    = {Yu, Xinhui and Wang, Dan
               and McKeown, Martin J. and Wang, Z. Jane},
  journal   = {IEEE Transactions on Artificial Intelligence},
  volume    = {5},
  number    = {4},
  pages     = {1525--1532},
  year      = {2024},
  doi       = {10.1109/TAI.2023.3298297}
}

@inproceedings{he2016deep,
  title     = {Deep Residual Learning for Image Recognition},
  author    = {He, Kaiming and Zhang, Xiangyu and Ren, Shaoqing
               and Sun, Jian},
  booktitle = {Proceedings of the IEEE Conference on Computer Vision
               and Pattern Recognition},
  pages     = {770--778},
  year      = {2016},
  doi       = {10.1109/CVPR.2016.90}
}

@inproceedings{loshchilov2019decoupled,
  title     = {Decoupled Weight Decay Regularization},
  author    = {Loshchilov, Ilya and Hutter, Frank},
  booktitle = {International Conference on Learning Representations},
  year      = {2019},
  url       = {https://openreview.net/forum?id=Bkg6RiCqY7}
}

@article{russakovsky2015imagenet,
  title     = {ImageNet Large Scale Visual Recognition Challenge},
  author    = {Russakovsky, Olga and Deng, Jia and Su, Hao
               and Krause, Jonathan and Satheesh, Sanjeev
               and Ma, Sean and Huang, Zhiheng
               and Karpathy, Andrej and Khosla, Aditya
               and Bernstein, Michael and Berg, Alexander C.
               and Fei-Fei, Li},
  journal   = {International Journal of Computer Vision},
  volume    = {115},
  number    = {3},
  pages     = {211--252},
  year      = {2015},
  doi       = {10.1007/s11263-015-0816-y}
}

@inproceedings{vaswani2017attention,
  title     = {Attention Is All You Need},
  author    = {Vaswani, Ashish and Shazeer, Noam
               and Parmar, Niki and Uszkoreit, Jakob
               and Jones, Llion and Gomez, Aidan N.
               and Kaiser, Lukasz and Polosukhin, Illia},
  booktitle = {Advances in Neural Information Processing Systems},
  volume    = {30},
  year      = {2017}
}

@inproceedings{dosovitskiy2017carla,
  title     = {{CARLA}: An Open Urban Driving Simulator},
  author    = {Dosovitskiy, Alexey and Ros, German
               and Codevilla, Felipe and Lopez, Antonio
               and Koltun, Vladlen},
  booktitle = {Proceedings of the 1st Annual Conference on Robot Learning},
  series    = {Proceedings of Machine Learning Research},
  volume    = {78},
  pages     = {1--16},
  year      = {2017},
  publisher = {PMLR},
  url       = {https://proceedings.mlr.press/v78/dosovitskiy17a.html}
}

@inproceedings{jia2024bench2drive,
  title     = {Bench2Drive: Towards Multi-Ability Benchmarking of Closed-Loop End-to-End Autonomous Driving},
  author    = {Jia, Xiaosong and Yang, Zhenjie
               and Li, Qifeng and Zhang, Zhiyuan
               and Yan, Junchi},
  booktitle = {Advances in Neural Information Processing Systems},
  year      = {2024},
  doi       = {10.52202/079017-0025}
}

@inproceedings{ioffe2015batch,
  title     = {Batch Normalization: Accelerating Deep Network Training by Reducing Internal Covariate Shift},
  author    = {Ioffe, Sergey and Szegedy, Christian},
  booktitle = {Proceedings of the 32nd International Conference on Machine Learning},
  pages     = {448--456},
  volume    = {37},
  series    = {Proceedings of Machine Learning Research},
  year      = {2015},
  publisher = {PMLR},
  url       = {https://proceedings.mlr.press/v37/ioffe15.html}
}

@article{chen2024end,
  title     = {End-to-End Autonomous Driving: Challenges and Frontiers},
  author    = {Chen, Li and Wu, Penghao and Chitta, Kashyap
               and Jaeger, Bernhard and Geiger, Andreas
               and Li, Hongyang},
  journal   = {IEEE Transactions on Pattern Analysis and Machine Intelligence},
  volume    = {46},
  number    = {12},
  pages     = {10164--10183},
  year      = {2024},
  doi       = {10.1109/TPAMI.2024.3435937}
}

@inproceedings{codevilla2018end,
  title     = {End-to-End Driving via Conditional Imitation Learning},
  author    = {Codevilla, Felipe and Muller, Matthias
               and Lopez, Antonio and Koltun, Vladlen
               and Dosovitskiy, Alexey},
  booktitle = {2018 IEEE International Conference on Robotics and Automation (ICRA)},
  pages     = {4693--4700},
  year      = {2018},
  doi       = {10.1109/ICRA.2018.8460487}
}

@inproceedings{zhong2023making,
  title={Making Batch Normalization Great in Federated Deep Learning},
  author={Zhong, Jike and Chen, Hong-You and Chao, Wei-Lun},
  booktitle={International Conference on Learning Representations},
  year={2023},
  url={https://arxiv.org/abs/2303.06530}
}

@inproceedings{li2021fedbn,
  title={FedBN: Federated Learning on Non-IID Features via Local Batch Normalization},
  author={Li, Xiaoxiao and Jiang, Meirui and Zhang, Xiaofei and Kamp, Michael and Dou, Qi},
  booktitle={International Conference on Learning Representations},
  year={2021},
  url={https://arxiv.org/abs/2102.07623}
}

@article{schonemann1966generalized,
  title={A Generalized Solution of the Orthogonal
         Procrustes Problem},
  author={Sch{\"o}nemann, Peter H.},
  journal={Psychometrika},
  volume={31},
  number={1},
  pages={1--10},
  year={1966},
  doi={10.1007/BF02289451}
}

@book{gower2004procrustes,
  title={Procrustes Problems},
  author={Gower, John C. and Dijksterhuis, Garmt B.},
  year={2004},
  publisher={Oxford University Press},
  doi={10.1093/acprof:oso/9780198510581.001.0001}
}

@inproceedings{achara2026multiway,
  title={Multi-Way Representation Alignment},
  author={Achara, Akshit and Gaintseva, Tatiana
    and Mahaut, Mat{\'e}o and Chakraborty, Pritish
    and Johansson, Viktor Stenby and Barsbey, Melih
    and Rodol{\`a}, Emanuele and Crisostomi, Donato},
  booktitle={Proceedings of the 43rd International
    Conference on Machine Learning},
  volume={306},
  pages={310--331},
  year={2026},
  publisher={PMLR}
}

@inproceedings{tung2019similarity,
  title={Similarity-Preserving Knowledge Distillation},
  author={Tung, Frederick and Mori, Greg},
  booktitle={Proceedings of the IEEE/CVF International Conference
             on Computer Vision (ICCV)},
  pages={1365--1374},
  year={2019},
  doi={10.1109/ICCV.2019.00145}
}

@inproceedings{chen2021cross,
  title={Cross-Layer Distillation with Semantic Calibration},
  author={Chen, Defang and Mei, Jian-Ping and Zhang, Yuan
          and Wang, Can and Wang, Zhe and Feng, Yan
          and Chen, Chun},
  booktitle={Proceedings of the AAAI Conference on
             Artificial Intelligence},
  volume={35},
  number={8},
  pages={7028--7036},
  year={2021},
  doi={10.1609/aaai.v35i8.16865}
}

@article{liu2020adaptive,
  title={Adaptive Multi-Teacher Multi-Level Knowledge Distillation},
  author={Liu, Yuang and Zhang, Wei and Wang, Jun},
  journal={Neurocomputing},
  volume={415},
  pages={106--113},
  year={2020},
  doi={10.1016/j.neucom.2020.07.048}
}

@inproceedings{iordache2025multilevel,
  title={Multi-Level Feature Distillation of Joint Teachers
         Trained on Distinct Image Datasets},
  author={Iordache, Adrian and Alexe, Bogdan
          and Ionescu, Radu Tudor},
  booktitle={Proceedings of the Winter Conference on
             Applications of Computer Vision (WACV)},
  pages={7133--7142},
  year={2025}
}

@article{gong2024bilevel,
  title={Bi-Level Orthogonal Multi-Teacher Distillation},
  author={Gong, Shuyue and Wen, Weigang},
  journal={Electronics},
  volume={13},
  number={16},
  pages={3345},
  year={2024},
  doi={10.3390/electronics13163345}
}

@inproceedings{zhang2021endtoend,
  title={End-to-End Federated Learning for Autonomous Driving Vehicles},
  author={Zhang, Hongyi and Bosch, Jan and Olsson, Helena Holmstr{\"o}m},
  booktitle={2021 International Joint Conference on Neural Networks (IJCNN)},
  pages={1--8},
  year={2021},
  publisher={IEEE},
  doi={10.1109/IJCNN52387.2021.9533808}
}

@inproceedings{zhang2021realtime,
  title={Real-time End-to-End Federated Learning: An Automotive Case Study},
  author={Zhang, Hongyi and Bosch, Jan and Olsson, Helena Holmstr{\"o}m},
  booktitle={2021 IEEE 45th Annual Computers, Software, and Applications Conference (COMPSAC)},
  pages={459--468},
  year={2021},
  publisher={IEEE},
  doi={10.1109/COMPSAC51774.2021.00070}
}

@inproceedings{aparna2021steering,
  title={Steering Angle Prediction for Autonomous Driving using Federated Learning: The Impact of Vehicle-To-Everything Communication},
  author={{Aparna M. P.} and Gandhiraj, R. and Panda, Manoj},
  booktitle={2021 12th International Conference on Computing Communication and Networking Technologies (ICCCNT)},
  pages={1--7},
  year={2021},
  publisher={IEEE},
  doi={10.1109/ICCCNT51525.2021.9580097}
}

@inproceedings{nguyen2022deep,
  title={Deep Federated Learning for Autonomous Driving},
  author={Nguyen, Anh and Do, Tuong and Tran, Minh and Nguyen, Binh X. and Duong, Chien and Phan, Tu and Tjiputra, Erman and Tran, Quang D.},
  booktitle={2022 IEEE Intelligent Vehicles Symposium (IV)},
  pages={1824--1830},
  year={2022},
  publisher={IEEE},
  doi={10.1109/IV51971.2022.9827020}
}

@inproceedings{kou2023communication,
  title={Communication Resources Constrained Hierarchical Federated Learning for End-to-End Autonomous Driving},
  author={Kou, Wei-Bin and Wang, Shuai and Zhu, Guangxu and Luo, Bin and Chen, Yingxian and Ng, Derrick Wing Kwan and Wu, Yik-Chung},
  booktitle={2023 IEEE/RSJ International Conference on Intelligent Robots and Systems (IROS)},
  pages={9383--9390},
  year={2023},
  publisher={IEEE},
  doi={10.1109/IROS55552.2023.10342134}
}

@inproceedings{lin2020ensemble,
  title={Ensemble Distillation for Robust Model Fusion
         in Federated Learning},
  author={Lin, Tao and Kong, Lingjing and Stich, Sebastian U.
          and Jaggi, Martin},
  booktitle={Advances in Neural Information Processing Systems},
  volume={33},
  pages={2351--2363},
  year={2020},
  publisher={Curran Associates, Inc.},
  url={https://proceedings.neurips.cc/paper/2020/hash/18df51b97ccd68128e994804f3eccc87-Abstract.html}
}

@inproceedings{jia2024bench,
  title={{Bench2Drive}: Towards Multi-Ability Benchmarking
         of Closed-Loop End-to-End Autonomous Driving},
  author={Jia, Xiaosong and Yang, Zhenjie and Li, Qifeng
          and Zhang, Zhiyuan and Yan, Junchi},
  booktitle={Advances in Neural Information Processing Systems},
  year={2024},
  doi={10.52202/079017-0025},
  url={https://mlanthology.org/neurips/2024/jia2024neurips-bench2drive/}
}

@inproceedings{xiao2023scaling,
  title={Scaling Vision-Based End-to-End Autonomous Driving
         with Multi-View Attention Learning},
  author={Xiao, Yi and Codevilla, Felipe and Porres, Diego
          and L{\'o}pez, Antonio M.},
  booktitle={2023 IEEE/RSJ International Conference on
             Intelligent Robots and Systems (IROS)},
  year={2023},
  publisher={IEEE},
  doi={10.1109/IROS55552.2023.10341506},
  url={https://doi.org/10.1109/IROS55552.2023.10341506}
}

@inproceedings{li2020federated,
  title={Federated Optimization in Heterogeneous Networks},
  author={Li, Tian and Sahu, Anit Kumar and Zaheer, Manzil
          and Sanjabi, Maziar and Talwalkar, Ameet
          and Smith, Virginia},
  booktitle={Proceedings of Machine Learning and Systems},
  volume={2},
  year={2020},
  url={https://proceedings.mlsys.org/paper/2020/hash/1f5fe83998a09396ebe6477d9475ba0c-Abstract.html}
}
}

\end{document}